\documentclass{article}

\usepackage{graphicx}
\usepackage{algorithm}
\usepackage{algorithmic}

\usepackage{subcaption}
\usepackage{amssymb}
\usepackage[export]{adjustbox}
\usepackage{subcaption,graphicx}

\usepackage[nonatbib, preprint]{neurips_2024}

\usepackage[utf8]{inputenc} 
\usepackage[T1]{fontenc}    
\usepackage{hyperref}       
\usepackage{url}            
\usepackage{booktabs}       
\usepackage{amsfonts}       
\usepackage{nicefrac}       
\usepackage{microtype}      
\usepackage{xcolor}         

\title{Natural Learning}

\author{%
  Hadi Fanaee-T\\
  School of Information Technology\\
  Halmstad University\\
  Box 823, 301 18 Halmstad, Sweden \\
  \texttt{hadi.fanaee@hh.se} \\
}

\begin{document}

\maketitle

\begin{abstract}
	
We introduce Natural Learning (NL), a novel algorithm that elevates the explainability and interpretability of machine learning to an extreme level. NL simplifies decisions into intuitive rules, like "We rejected your loan because your income, employment status, and age collectively resemble a rejected prototype more than an accepted prototype." When applied to real-life datasets, NL produces impressive results. For example, in a colon cancer dataset with 1545 patients and 10935 genes, NL achieves 98.1\% accuracy, comparable to DNNs and RF, by analyzing just 3 genes of test samples against 2 discovered prototypes. Similarly, in the UCI's WDBC dataset, NL achieves 98.3\% accuracy using only 7 features and 2 prototypes. Even on the MNIST dataset (0 vs. 1), NL achieves 99.5\% accuracy with only 3 pixels from 2 prototype images. NL is inspired by prototype theory, an old concept in cognitive psychology suggesting that people learn single sparse prototypes to categorize objects. Leveraging this relaxed assumption, we redesign Support Vector Machines (SVM), replacing its mathematical formulation with a fully nearest-neighbor-based solution, and to address the curse of dimensionality, we utilize locality-sensitive hashing. Following theory's generalizability principle, we propose a recursive method to prune non-core features. As a result, NL efficiently discovers the sparsest prototypes in $O(n^2pL)$ with high parallelization capacity in terms of n. Evaluation of NL with 17 benchmark datasets shows its significant outperformance compared to decision trees and logistic regression, two methods widely favored in healthcare for their interpretability. Moreover, NL achieves performance comparable to finetuned black-box models such as deep neural networks and random forests in 40\% of cases, with only a 1-2\% lower average accuracy. The code is available via \url{http://natural-learning.cc}.

\end{abstract}

\section{Introduction}                                                                                                                                                            Black-box models such as Deep Neural Networks (DNNs) have achieved state-of-the-art performance on benchmark datasets and real-world applications. However, these models operate as black-box systems, unable to explain their predictions in a manner understandable to humans.

A typical practice to alleviate problems of black-box models is to construct a secondary (post-hoc) model to explain the original model (e.g., SHAP \cite{lundberg2017unified} or LIME \cite{ribeiro2016should}). However, as Rudin \cite{rudin2019stop} argues, \textbf{post-hoc explanations are inherently wrong}. They cannot achieve perfect fidelity concerning the original model. If the explanation were entirely faithful to the original model, it would essentially replicate the original model itself, rendering the need for the original model redundant. 

Label noise is a prevalent issue in datasets. It has been reported that real-world datasets typically contain between 8.0\% and 38.5\%  label noise \cite{semenova2024path,song2022learning,song2019selfie,lee2018cleannet}. Recently, \cite{semenova2024path} provided theoretical evidence that in noisy datasets (such as datasets about humans like healthcare), \textbf{simple, interpretable classifiers should perform as well as black-box models}.

Simple models like logistic regression offer an excellent level of interpretability and explainability. Nevertheless, linear models lack a built-in mechanism to deal with \textbf{irrelevant features}, the \textbf{curse of dimensionality}, and multicollinearity, all of which can result in overfitting or decreased model performance, particularly in high-dimensional datasets.

Decision trees \cite{quinlan1986induction} are a more straightforward option for understanding how a model makes decisions. Compared to linear methods, they better handle high-dimensional data, irrelevant features, and noisy samples. However, while feature importance inferred from decision trees may suggest they are explainable, it's not guaranteed that all decisions are based on those essential features. Unlike logistic regression, decision trees can't explain decisions broadly. They mainly provide local explanations, which may be useless if we can't generalize them. Moreover, interpreting decision trees becomes challenging as the tree's depth increases, leading to many terminal nodes that are difficult for humans to process \cite{quinlan1986induction}. Additionally, each decision may involve different feature combinations, making it hard to understand \textbf{interactions between features} at a global level. 

The challenges associated with decision trees stem from their underlying philosophy, which is rooted in Aristotle's categorization theory \cite{ryle1937categories}. This theory assumes that individuals employ rule-based explanations to categorize concepts. However, research in cognitive psychology has indicated shortcomings in this model, suggesting that\textbf{ people likely do not rely on rule-based definitions when categorizing objects}. According to Rosch's prototype theory \cite{rosch1973natural}, \textbf{people instead categorize objects by their perceived similarity to a prototypical (or ideal) example} of the category. 

Regarding interpretable prototype-based learning, we have almost no available options. The most related solutions are prototype selection \cite{bien2012prototype},  Nearest centroid classifier \cite{tibshirani2002diagnosis}, and Support Vector Machines (SVM) \cite{boser1992training}. However, \textbf{none of these methods meet the conditions of prototype theory}, mainly in two aspects: 1) \textbf{do not limit the number of prototypes for each class to one}; 2) \textbf{They are not equipped with a mechanism to impose sparsity on the features of prototypes, as theory suggests}. Hence, they cannot be a resemblance to the prototype theory. This issue represents a \textbf{significant gap} that drives all needs for explainability and interpretability towards decision trees, which, as we argued earlier, have substantial shortcomings.

We introduce \textbf{the first interpretable prototype-based classifier}, called "Natural Learning (NL)," which is the closest machine replica of prototype theory. NL has some attractive properties that make it distinct comparing other learning paradigms: 1) it stands as the sole instance-based classifier that \textbf{doesn't estimate any weights or rules} and despite being \textbf{hyperparameter-free}, \textbf{does not memorize the entire training set}; 2) it is the only instance-based classifier that offers explainability and interpretability in the form of human-like reasoning \cite{langley2022computational}: "If X is closer to A's prototype comparing B's prototype it is labeled A, otherwise, B."); 3) it has the \textbf{lowest model variance} among all classifiers, even lower than DNNs; 4) Its models are the \textbf{sparsest among all classifiers}, and consequently, offers the \textbf{highest prediction speed}.

\section{Methodology}

According to prototype theory \cite{rosch1973natural}, people categorize objects and concepts based on their similarity to a prototype. For instance, when we decide whether to classify a sofa into either the furniture category or the electronics category, we look at the similarity of the sofa with prototypes of those categories and assign the object to the nearer category. Prototypes have four primary characteristics: 1) they are the most typical examples of a category (\textbf{typicality principle}). For instance, a basic chair serves as the prototype for the furniture category; 2) prototypes are identified by their core features (\textbf{sparsity principle}), which are the central features of the prototype that are typically shared by most, if not all, instances within the category and are necessary for distinguishing the category from other categories. For example, we identify a basic chair by its seat and legs; 3) features of a prototype are generalizable to other members of the category (\textbf{generalizability principle}), even if those members differ in some respects from the prototype itself. For example, all members of the chair category should possess a seat and legs to be classified as chair; and 4) prototypes are not fixed entities; they can change based on new experiences (\textbf{flexibility principle}).

In the machine learning context, we can translate the properties mentioned above as follows: 1) A \textbf{single support vector} represents each class; 2) The features of support vectors are \textbf{sparse}; 3) Features of support vectors are \textbf{generalizable}; 4) Learning sparse support vectors is not a single-step process; instead, it is an \textbf{incremental} process. These are all the information needed to import prototype theory into machine learning in its original form. However, as mentioned earlier, no machine learning algorithm simultaneously implements these properties.

As the initial attempt to implement prototype theory, we can frame it as a pure cross-validation problem. We evaluate all possible pairs of samples from classes 0 and 1, along with all subsets of features (from length 1 to p), to determine which one generalizes well to all samples. Subsequently, we select the one with the lowest generalization error and the fewest features. As shown in the Appendix, when we apply this straightforward method to the iris dataset with an 80/20 train/test split (seed=42), we achieve a test accuracy of 100\% for Setosa vs. Versicolour. This results in a simple classification rule: If the petal length of the test sample is closer to sample 5's petal length (1.4) than sample 6's petal length (3.3), the prediction is Setosa (0); otherwise, it is Versicolour (1). In this rule, sample 5 is the prototype of the Setosa class, and sample 6 is the prototype of the Versicolour class found by the algorithm. Surprisingly, no one has attempted this simple classifier before. This model is both interpretable and explainable: the rule refers to two exact prototype cases in the training set. Moreover, this method is inherently robust in labeling noise (noisy samples) and highly sparse. We only need to store one feature of two training samples to make 100\% accurate predictions, and we no longer need the training set. Therefore, it exhibits attractive properties not seen before in any other classifier.

When considering the practicality of solutions, this naive cross-validation method falls short. For an $n \times p$ feature matrix, evaluating the generalization of prototype candidates (pairs of examples from + and - classes) and their corresponding features requires $O(n^32^p)$ time. The $2^p$ factor is dominant in typical datasets, rendering it infeasible. Additionally, this method is susceptible to the curse of dimensionality, as nearest neighbors lose meaning in higher dimensions. Moreover, this method is not robust enough to handle noisy features. Unfortunately, prototype theory doesn't offer solutions to these problems. These issues only arise in computational contexts and necessitate computing solutions. Therefore, we need to develop a more efficient method that is scalable and robust against the curse of dimensionality and noisy features.

To address the problem of the curse of dimensionality, we should explore nature's solution for nearest neighbor search. A recent study on the olfactory circuit of the fruit fly \cite{dasgupta2017neural} reveals that the fruit fly's brain employs a specialized version of the Locality-Sensitive Hashing (LSH) algorithm \cite{indyk1998approximate} for nearest neighbor search. Therefore, it appears that nature's mechanism for circumventing the curse of dimensionality and enabling efficient nearest-neighbor search involves an LSH-based method, which could quickly solve our first problem.

Since prototype theory suggests that we have only one support vector from each class, \textbf{revisiting soft-margin SVM} \cite{cortes1995support} can offer new insights. Suppose A is a margin violation sample and (B, C) are support vectors for the positive and negative classes. The nearest neighbor of A (a positive example) from its class is B (the support vector of the positive class), and its closest neighbor from the opposite class is C (the support vector of the negative class). By testing all samples, including A,\textbf{ it is guaranteed that at least in 1 out of n validations, we find support vectors B and C without conducting any expensive optimization}. Since B and C are actual support vectors, their generalizability testing surpasses that of any pair in the training set. This approach eliminates the need to test all possible pairs in the training set and reduces the complexity of $n$ from $O(n^2)$ to $O(n)$. It's important to note that this solution only becomes viable due to the relaxed assumption of prototype theory (typicality principle), perhaps explaining why no one has attempted this before.

Margin violation samples not only serve as a natural guide to finding singular support vectors, but they can also contribute to addressing the second and third problems (noisy features and complexity concerning $p$). If we compare the individual feature values of A to B and C, those features that make A seem closer to C (the class that A does not belong to) are likely non-core features and can be considered for pruning. Removing them and achieving a better generalization error indicates that they were indeed non-core features that negatively affected our classification accuracy. \textbf{After removing non-core features, we enter a new feature space with a more decisive contribution of core features}. This results in an improved nearest-neighbor search and purer classes. If we repeat this process recursively, we will eventually pinpoint the core features of prototypes that are so robust that they can't be pruned further. This method prunes non-core features and reduces the complexity of $p$ from $O(2^p)$  to $O(pL)$, where L is the number of recursive iterations.

\section{Natural Learning}
This section gives a detailed look at our new algorithm, which combines ideas from earlier discussions in a simple and unified way. Natural Learning is math-free and optimization-free. It simply applies the principles of prototype theory with great precision.

\subsection{Definitions}

Let $ (\mathbf{x}_1, y_1), (\mathbf{x}_2, y_2), \ldots, (\mathbf{x}_n, y_n) $ be the training dataset, where $x_i$ is the $ i $-th training instance and $D = \{1, 2, \ldots, p\}$ is the indices of features; $ y_i $ is the corresponding class label ($ y_i = \{0, 1\} $) for two-groups classification).

Let denote the index of sample $i$'s nearest neighbor of the same-class and opposite-class with $s$ and $o$, respectively. Then, the absolute value of element-wise subtraction of $x_i$ from $x_s$ and $x_o$ can be shown with $v_s=|\mathbf{x}_s-\mathbf{x}_i|_j$ and $v_o=|\mathbf{x}_o-\mathbf{x}_i|_j$ for $j=1,2,...,p$. Let denote element-wise subtraction of $v_o$ from $v_s$ as $v=(v_o-v_s)_j$ for $j=1,2,...,p$. Indices of prototype features is then defined as $C_i=\{j|v_j>0\}$ where j represents the index of elements in the resulting vector $v$ such that the value of the element at index j is greater than zero. 
Then, prototype sample candidates are $\mathbf({z}_s, y_s)$ and $\mathbf({z}_o, ~y_o)$, where $z_i = (x_{i1}, x_{i2}, ..., x_{ip})_{C_i}$.

\subsection{Training Algorithm}

\begin{algorithm}[htbp]
	\small
	\caption{NLTrain}
	\label{alg_NLTrain}
	\begin{algorithmic}[1]
		\STATE Input: training set $(x,y)$ ($n$ samples and $p$ features), $ y_i = \{0, 1\} $, and features of best prototype $M$

		\STATE Output: prototype samples ($s_{best}$ and $o_{best}$), and their labels, prototype features  $M$
		\IF{$M$ is null}
		\STATE $M \gets \{1,2,...,p\}$ \hspace{1cm} \textit{//initialization
		 	 of prototype features} 
	 	\ENDIF
	 	\STATE $x=x(:,M)$ \hspace{1cm} \textit{// Copy of $x$ with features in $M$} 
		\STATE $e_{best} \gets \infty$ \hspace{1cm} \textit{//initialization of best error. Allowing NL to learn better prototypes at each iteration.} 
		\FOR{each sample $i$ in $x$}
		\STATE $s \gets$ index of $x_i$'s nearest neighbor from same class using LSH \hspace{0.5cm} \textit{//prototype sample candidate}
		\STATE $o \gets$ index of  $x_i$'s nearest neighbor from opposite class using LSH \hspace{0.5cm} \textit{//prototype sample candidate}
		\STATE $C \gets$ indices of features in $M$ that make $x_i$ closer to $x_s$ than $x_o$ \hspace{0.5cm}\textit{// prototype features candidate}
		\STATE $\hat{y} \gets NLPredict(x_s, x_o, y_s,y_o, C, x)$ \hspace{0.2cm} \textit{// test the generalization of prototype candidate} 
		\STATE $e \gets \sum(y \neq \hat{y})$
		\IF {$e<e_{best}$ \& $|C|>1$}
		\STATE $(s_{best},o_{best}) \gets (s,o)$    \hspace{1cm} \textit{// Best prototype samples} 
		\STATE $C_{best} \gets C$    \hspace{1cm} \textit{//Best prototype features} 
		\STATE $e_{best} \gets e$ \hspace{1cm} \textit{//Best error so far} 
		\ENDIF
		\ENDFOR
		\IF {$|C_{best}|\neq|M|$} 
		\STATE $M \gets C_{best}$
			\STATE $NLTrain(x,y,M)$
		\ENDIF

	\end{algorithmic}
\end{algorithm}

\begin{algorithm}[htbp]
	\small
	\caption{NLPredict}\label{alg_NLPredict}
	\label{alg:algorithm1}
	\begin{algorithmic}[1]
		\STATE Input: data (x), prototype samples ($x_o$ and $x_s$) and corresponding labels ($y_o$ and $y_s$) and  features ($M$)
		\STATE Output: $\hat{y}$ (Predicted labels) 
		\STATE $x \gets x(:,M)$  \hspace{1cm}// copy of $x$ with prototype features ($M$ or $C_i$)
		\FOR{each sample $i$ in $x$}
		\STATE $(d_s,d_o) \gets D(x_i, x_s, x_o)$ \hspace{1cm} \textit{//Distance of example to prototype samples s and o} 
		\STATE $\hat{y_i}=y_s$ 
		\IF{$d_o<d_s$}
		\STATE $\hat{y_i}=y_o$ \hspace{1cm} 
		\ENDIF
		\ENDFOR
		
	\end{algorithmic}
\end{algorithm}

The training algorithm is simple and intuitive. It comprises basic operations such as LSH-based nearest neighbor search and element-wise subtraction of vectors. As shown in Algorithm \ref{alg_NLTrain}, we start by initializing features of prototypes with indices of all features. Then, we create a copy of the feature matrix, keeping only prototype features. Later, for the sample $x_i$, we form a local triplet consisting of $x_i$ and its nearest neighbor from its class ($x_s$) and the opposite class ($x_o$). Next, through element-wise subtraction of feature values, we compare $x_i$ with $x_s$ and $x_o$ to prune features that make $x_i$ closer to $x_o$. This task is equivalent to the elimination of non-core features of the prototype candidate. The obtained prototype candidates and their reduced feature subset from the local triplet become our candidate for prototype samples/features. If the number of features is more than one, we check how well this decision boundary generalizes to all instances by computing the global generalization error ($e$). We pick the candidate that offers the lowest error. The features of the winning prototype serve as the new global subset of features. We recursively repeat the process until selected features for the best prototype (core features) become so robust that they don't change further.

The cross-validation or prediction process is depicted in Algorithm \ref{alg_NLPredict}. To execute it, we need to input a vector containing prototype vectors and their respective labels, along with the index of prototype features. Using the designated prototype features, we then calculate the similarity between unseen and prototype examples. We assign the label of the closest prototype to the unseen example.

\subsubsection{Illustrative Example}

\begin{figure}[htbp]
	\centering
	\includegraphics[width=0.9\textwidth]{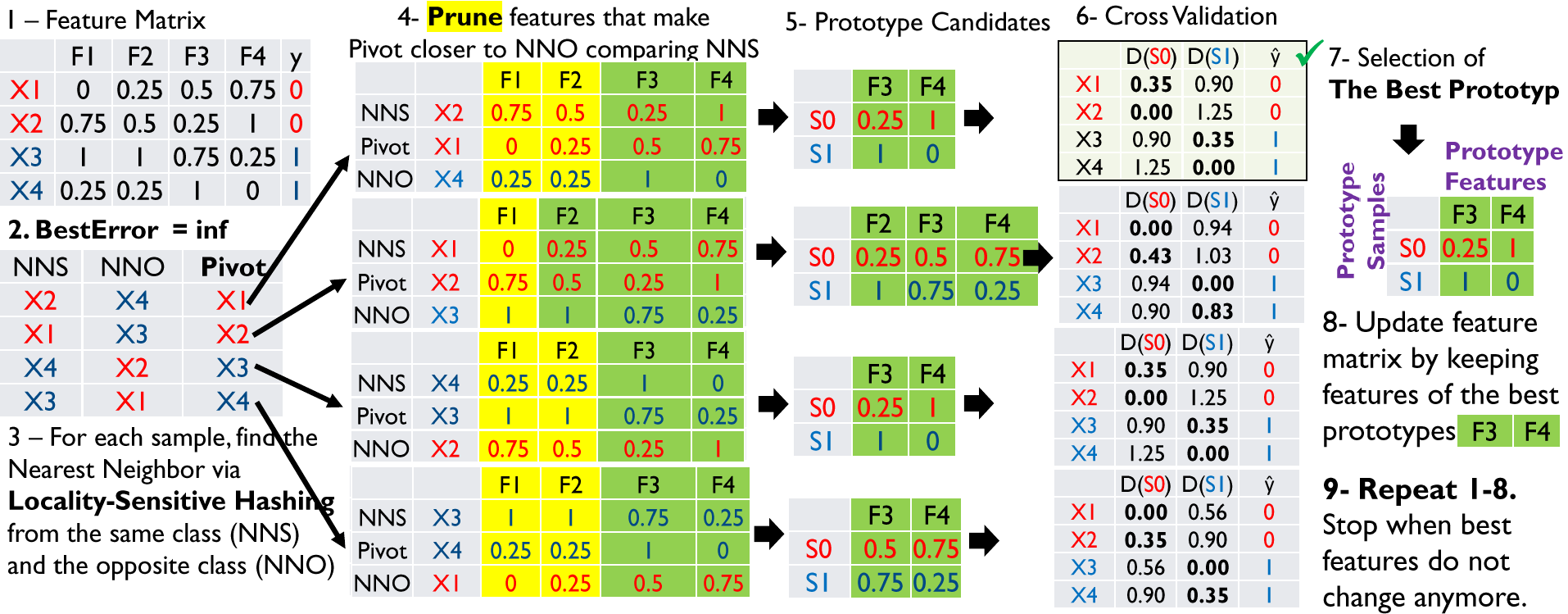}
	\caption{NL's first training iteration on a toy dataset. Step 3 nominates candidates for support vectors, and step 6 tests their typicality; Step 4 implements the principle of "generalizability of features" from prototype theory; Steps 7-9 mimic the "flexibility principle" of prototype theory, allowing NL to switch better sparser prototypes at each iteration.}\label{fig_toyexample}
\end{figure}

Fig. \ref{fig_toyexample} shows an illustrative example of our training algorithm on a toy dataset of four samples and four features. We first form a triplet of training sample (Pivot), the nearest neighbor from its class (NN-S), and the closest neighbor from the opposite class (NN-O). Then, in step 4, we construct a feature comparison table for each pivot to identify candidate features for pruning. Let's take feature F1 of the first comparison table as an example. F1's value for the pivot and NN-S and NN-O are 0, 0.75, and 0.25, respectively. As we can see, feature F1 harms the pivot sample to appear closer to the opposite neighbor than its same-class neighbor. Thus, we prune F1. On the contrary, F3's values for pivot, NN-S, and NN-O are  1, 0.75, and 0, respectively. F3 does not negatively contribute to the pivot's similarity to its same-class neighbor. So, we keep it.

\begin{figure}[htbp]
	\centering
	\includegraphics[width=0.9\textwidth]{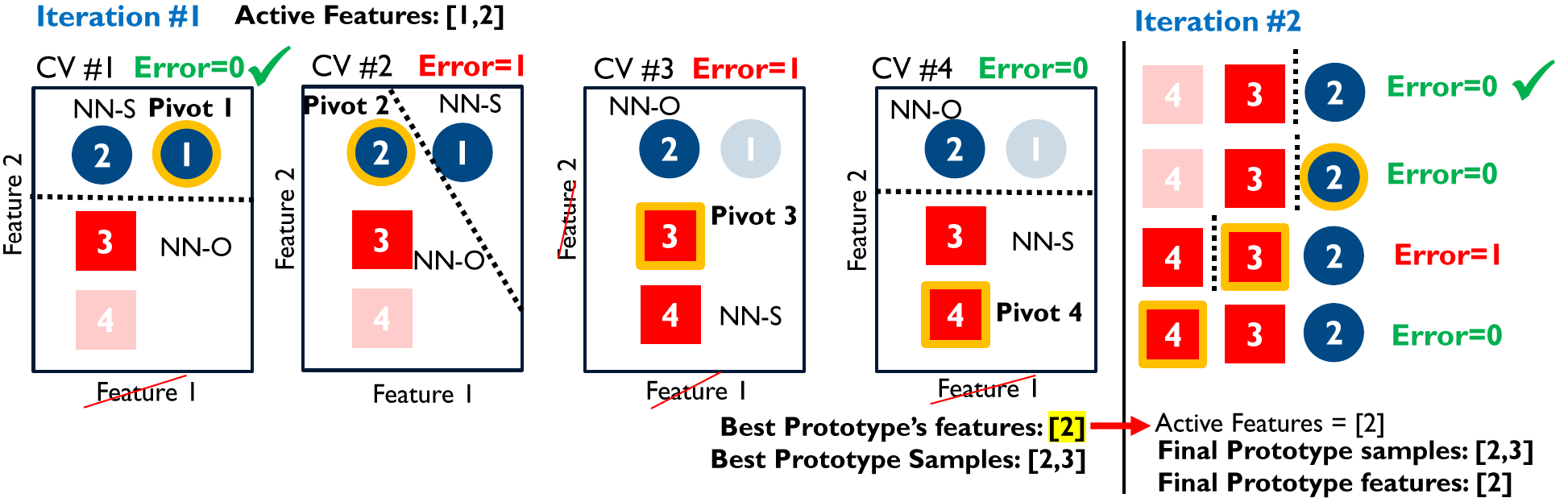}
	\caption{Geometric view of natural decision boundaries found by triplets in iterations 1 and 2.}\label{fig_geometricview}
\end{figure}

Fig. \ref{fig_geometricview} provides a geometric view of the algorithm's iterative process (with no connection with Fig. \ref{fig_toyexample}). We start with an active feature set, which at the beginning is the entire feature set (1,2). We establish the decision boundary of each triplet by calculating the distance of each example to the prototype candidates \{NN-S, NN-O\}. The example's label is the same label as the nearer prototype candidate. We observe that the decision boundary from two triplets results in zero prediction error for all training instances (CV1 and CV4). Feature 1 has an equal distance to both NN-O and NN-S. Therefore, we prune it (i.e., x-axis). Since this is the best error obtained, we pick samples 2 and 3 as prototype samples and feature two as the features of the prototypes. 

In the next iteration, the active feature set is defined as the features of the best-found prototypes (in this case, feature 2). Therefore, by removing feature 1, training samples are now represented as one-dimensional vectors with only feature 2. This change modifies the similarity of samples and their neighborhood structure. As we can see,  three candidates out of four have zero generalization errors. All of them point to samples 2 and 3 as prototype samples and feature 2 as the features of prototypes. As we can see, the composite of features does not change compared to the previous iteration; thus, we stop and introduce the best prototype samples and features as the model's output.

\section{Empirical Evaluation}
The objective of the empirical evaluation is to evaluate the position of NL concerning its main competitors: linear SVM (prototype-based), 1NN (nearest neighbor as the only hyperparameter-free, instance-based classifier), as well as decision trees and logistic regression, as the popular representatives of interpretable and explainable classifiers. We are also interested in evaluating the position of NL concerning state-of-the-art black-box models such as DNN, RF, and SVM with nonlinear kernels. 


\subsection{Datasets}

\begin{figure}[htbp]
	\centering
	\includegraphics[width=\textwidth]{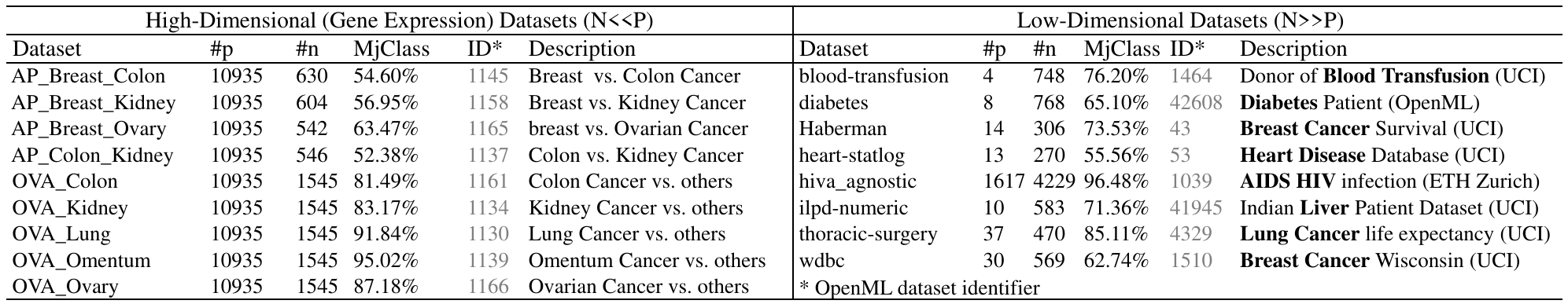}
	\caption{Public datasets from OpenML for binary classification.}\label{fig_datasets}
\end{figure}

We retrieve 17 benchmark datasets for binary classification from the OpenML data repository (See Fig. \ref{fig_datasets}), most of which are also available in UCI except for gene expression datasets. Almost half of the datasets are high-dimensional, while the other half are low-dimensional. We are particularly interested in the healthcare application domain, where NL would provide the highest value regarding interpretability and explainability. Another reason is that in this domain, we frequently encounter prototype patients, which aligns perfectly with the underlying assumption of NL. Therefore, all datasets are from this domain. To reduce bias in the train-test split, we generated ten stratified samples for each dataset (10-folds), resulting in a total of 170 datasets for training.

\subsection{Experimental Configuration}
We finetune baseline classifiers according to the following configurations. Decision trees: MaxSplits =[1, 5, 10, 20, 50, n], MinLeafSize=[1, 5, 10, 20, 50], in total 30 configurations. Linear SVM: C=[100, 10, 1.0, 0.1, 0.001], in total 5 configurations. SVM-RBF: same C values and gamma= [$2^{-16}:..:2^8$] as suggested by \cite{fernandez2014we} with step of $2^2$, in total 65 combinations. RF: same hyperparameter range for DT and NumTrees= [10,50,100], a total of 90 combinations. DNN: fixed batch size of 32, stochastic gradient descent as an optimizer, and max epoch of 20 plus grid search for Hidden Layers=[10, 30, 50], Layers=[2, 3, 4], Learning Rate=[0.01, 0.001] and Activation Functions=\{RelU, Tanh, Sigmoid\}, in total 54 combinations. We also normalize the data using min-max scaling for 1NN, LR, SVM, and DNN since their performance degrades without feature scaling.
 
\subsection{Evaluation Metrics}
We use accuracy and f-measure as our two main evaluation metrics. We use critical difference (CD) diagrams for statistical comparisons of classifiers over multiple data sets \cite{demvsar2006statistical}, which uses Nemenyi's test with alpha=0.01.

\section{Result}

\begin{figure}[htbp]
	\centering
	\includegraphics[width=1\textwidth]{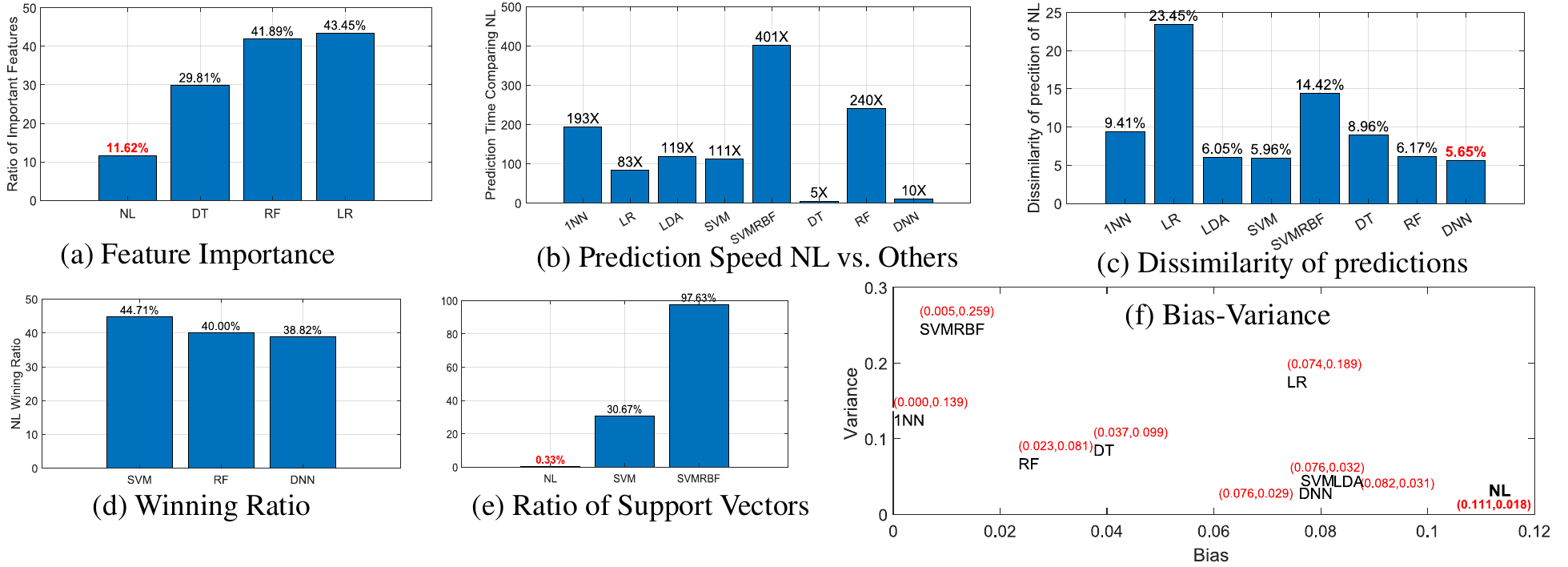}
	\caption{Performance of NL versus baseline methods in 170 stratified test sets (17 datasets).}\label{fig_results}
\end{figure}

\subsection{Accuracy and F-Measure: NL vs. others}
\begin{figure}[htbp]
	\centering
	\begin{subfigure}[b]{0.45\textwidth}
		\centering
		\includegraphics[width=\textwidth]{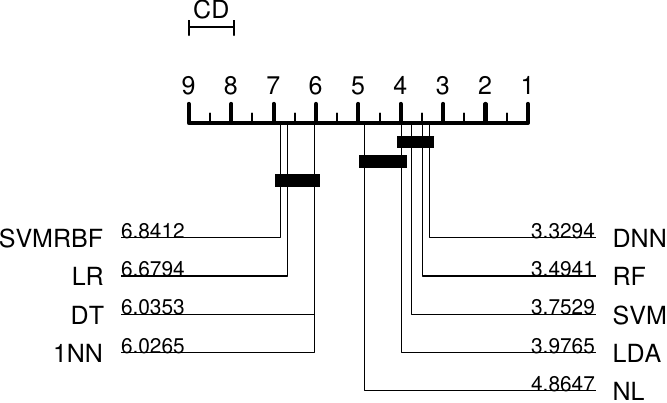}
		\caption{Accuracy}
		\label{fig:image2}
	\end{subfigure}
	\hfill 
	\begin{subfigure}[b]{0.42\textwidth}
		\centering
		\includegraphics[width=\textwidth]{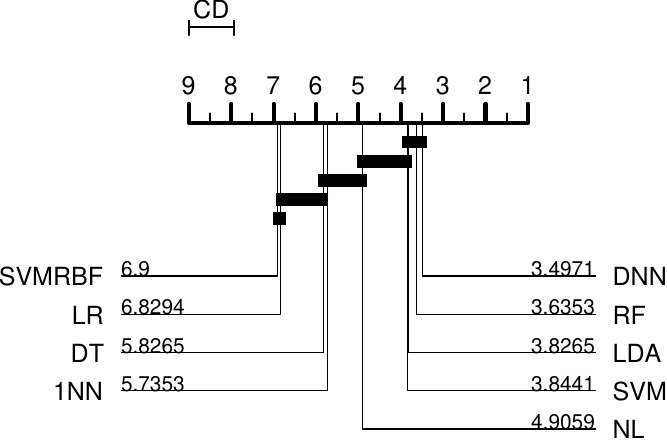}
		\caption{F-measure}
		\label{fig:image1}
	\end{subfigure}
	\caption{CDD of Accuracy and F-Measure. Horz. line implies no statistical significance at $\alpha=0.01$}
	\label{fig_perf}
\end{figure}

CD Diagram of accuracy and F-measure over 170 training sets (10 stratified samples of 17 datasets) is presented in Fig. \ref{fig_perf}. Regarding accuracy, as we can see, no statistically significant difference exists between NL and LDA. However, NL significantly outperforms decision trees ( average 1.5\% difference), logistic regression (average 13.5\% accuracy difference), and 1NN (in average almost 1.7\% accuracy difference), as well as SVM with RBF Kernel (average 14\% accuracy difference). However, DNN, RF, and SVM significantly outperform NL, respectively, with an average accuracy difference of 1-2\%. A surprisingly low performance is obtained for SVM with the RBF kernel. Generally, overfitting issues of kernel methods can justify this due to hyper-sensitivity to kernel parameters \cite{han2014overcome}. Regarding the f-measure, NL doesn't exhibit a statistical difference compared to SVM, LDA, 1NN, and DT.

\subsection{Scalability}
The worst-case time complexity of NL is $O(n^2p^2)$, where we need to iterate exactly $p$ times. However, NL is very aggressive in pruning the features. Therefore, most of the time, it comes up with reasonable prototypes with few iterations. For example, in the MNIST dataset (4 vs. 9), the most challenging pair, we find the prototypes in 6 iterations (n=11791). NL’s average complexity is $O(n^2pL)$, where L is usually a constant factor lower than 20. LSH can reduce the dominant factor, $n^2$, by early quitting when distant or irrelevant neighbors form the triplet. On the same MNIST dataset, LSH reduced 21\% of operations. A perfect property that makes NL scalable is that \textbf{it can be highly parallelized with GPUs}.

\subsection{Test Runtime: NL vs. others}
As illustrated in Fig. \ref{fig_results}-b, NL advances the frontier of the state-of-the-art regarding prediction speed due to its tiny model size: 1/5 of Decision Trees and 1/10 of DNN.

\subsection{Predictions' Explainability and Interpretability: NL vs. DT, RF, LR, and SVM}

We quantitatively evaluate the interpretability level of NL in comparison to its main competitors: SVM from an instance-based family, DT/RF from a tree-based family, and LR from linear models. We calculate the average ratio of prototype samples to the total number of samples (Fig. \ref{fig_results}-e) and the ratio of the prototype features to the total number of features (Fig. \ref{fig_results}-a). NL, on average, uses 12\% of features (prototype features) versus 30\% of pruned DT, 42\%  of pruned RF, and 43\%  of LR. NL is restricted to finding exactly two support vectors (in average 0.33\%  of samples), while linear SVM and SVM-RBF, on average, find 31\%  and 97\%  of samples as support vectors, respectively. This result shows that NL is more explainable and interpretable than its competitors. 

\subsection{Bias-Variance trade-off: NL vs. others}

Fig. \ref{fig_results}-f shows NL's average bias and variance compared to other classifiers. As can be seen, NL provides the lowest average variance among all classifiers. This extraordinarily low variance is related to the simplicity of the model and larger Rashomon ratio \cite{breiman2001statistical} due to noisy labels \cite{semenova2024path}.

\section{Examples of Discovered Sparse Prototypes by Natural Learning}

\begin{figure}[htbp]
	\centering
	\includegraphics[width=\textwidth]{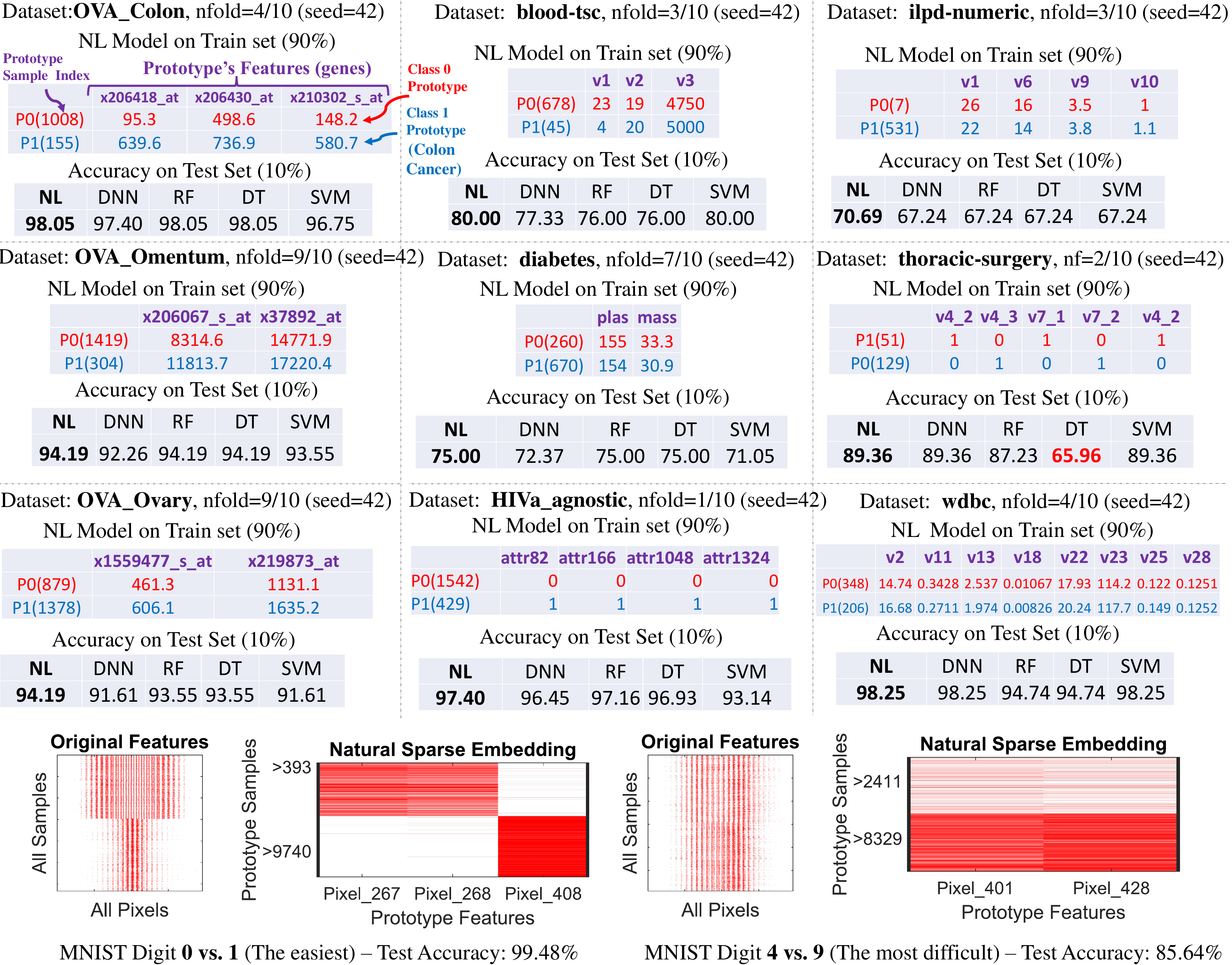}                      
	\caption{Example of NL's sparse models and their performance}\label{fig_examples}
\end{figure}

NL provides a surprising level of explainability and interpretability that has not been observed in any classifier before. As shown in Fig. \ref{fig_examples}, NL's models for nine datasets can fit into less than a half-page and are very human-friendly. For instance, the first case, OVA\_Colon, is a gene expression dataset. As the NL model suggests, to diagnose someone with colon cancer, we should pick patients 1008 and 155, focusing on three genes \textit{x206418\_at},\textit{ x206430\_at}, \textit{x210302\_s\_at}, and then a new case is diagnosed with colon cancer if her three gene values are closer to patient 155 than patient 1008. This simple NL model provides a prediction accuracy of 98.05\% on the test set, which is as good as RF and DT and better than SVM and DNN. Another example is the well-known UCI dataset of the Wisconsin Diagnostic of Breast Cancer (WDBC). NL has found seven prototype features and two patients, 348 and 206, as prototype samples. This NL model provides 98.25\% accuracy on the test set, equal to DNN and SVM and better than RF and DT.

We also conduct two experiments with MNIST, one on the most straightforward task, i.e., digit 0 vs. 1, and the other on the most difficult one: 4 vs. 9. As we can see, NL can find natural sparse embedding in both cases, but then the simple prototype rule has lower accuracy on the latter case that shows the limitation of NL. NL is not equipped with any representation learning mechanism. It entirely relies on original data representation and, consequently, cannot learn better representations.

\section{Conclusion and Future work}

Prototype theory has been recognized as a Copernican revolution in categorization theory because of its departure from the Aristotelian rule-based approach. Now, \textbf{we expect the same effect in machine learning}: a transition from decision trees towards natural learning, that provides much better human-like reasoning \cite{langley2022computational}. As we showed, NL can be more accurate than decision trees.

We anticipate that NL's sparse models provide new insights in many domains. Specifically, NL can be pivotal in various contexts: 1) In applications prioritizing \textbf{interpretability, explainability, and transparency}, where minor differences in accuracy are acceptable compared to black-box models, NL can \textbf{replace or complement decision trees and logistic regression} due to its simple \textbf{human-friendly and fair explanations}; 2) In applications where humans are the samples such as\textbf{ healthcare, criminal justice, and finance}, NL can provide a high value. In these domains, typically, labels are noisy, and black-box models provide the same performance as simple models \cite{semenova2024path,rudin2019stop}. Another reason is that in these applications, the existence of a prototype is guaranteed: \textbf{a clinical case in healthcare, a case study in finance, and a precedent case in criminal justice}; 3) In \textbf{discriminant analysis of high-dimensional omics data} (e.g., gene expression), NL can overcome the curse of dimensionality and the challenge of limited samples, and generate highly sparse and interpretable models that are essential in these domains. 4) NL models are very sparse both in sample and feature dimensions, making them suitable for \textbf{real-time applications} where \textbf{prediction speed is crucial} (e.g., defense, online trading) and for \textbf{embedded systems} where \textbf{processing and memory constraints} exist (e.g., wearable devices); 5) In handling \textbf{high-dimensional binary data} where dimensionality reduction or representation learning is not effective anymore, NL offers a promising alternative; 6) In the field of vision, NL does not appear to be competitive due to \textbf{lack of a mechanism for representation learning }. However, NL can still be helpful for \textbf{classification of trivial cases} (e.g., digit 0 vs. 1 in MNIST or frog vs. airplane in CIFAR-10 that we can get 86\%  accuracy) to gain a better prediction speed and explainability.
	
The essential open problems remain as future work: 1) \textbf{Is it possible to implement a local representation learning in NL's local triplet space}? We believe any meaningful results in this direction can result in the \textbf{white-box version of DNNs}. 2) Can ensemble methods boost NL's performance without harming its explainability? 3) Is it possible to extend NL for regression?

\bibliographystyle{plainnat}
\bibliography{bibliography}


\end{document}